# Integrating Artificial Intelligence Models and Synthetic Image Data for Enhanced Asset Inspection and Defect Identification

**[1]Reddy MANDATI**
**reddy.mandati@exeloncorp.com**

**[1]Vladyslav ANDERSON**
**vladyslav.anderson@exeloncorp.com**

**[1]Po-Chen CHEN***
**po-chen.chen@exeloncorp.com**

**[1]Ankush AGARWAL**
**ankush.agarwal@exeloncorp.com**

**[2]David BARNARD**
**david.barnard@bge.com**

**[2]Michael FINN**
**michael.finn@bge.com**

**[2]Jesse CROMER**
**jesse.cromer@bge.com**

**[1]Tatjana DOKIC**
**tatjana.dokic@exeloncorp.com**

**[2]Andrew MCCAULEY**
**andrew.mccauley@bge.com**

**[2]Clay TUTAJ**
**clay.tutaj@bge.com**

**[2]Neha DAVE**
**neha.dave@bge.com**

**[1]Bobby BESHARATI**
**bbesharati@pepcoholdings.com**

**[2]Jamie BARNETT**
**jamie.barnett@bge.com**

**[1]Timothy KRALL**
**timothy.krall@exeloncorp.com**

**[1]Exelon Corporation, USA, [2]BGE, An Exelon Company, USA**

**SUMMARY**

In the past utilities had to rely on in-field inspections to identify asset defects. This process is time-consuming and exposes crews to a variety of hazards. In addition, not all the defects could be captured by ground-based inspection. Recently, utilities have started using drone-based inspections to enhance the field-inspection process. This has resulted in utilities collecting and storing hundreds of thousands of images that require cataloguing and processing. We consider a vast repository of drone images, providing a wealth of information about asset health and potential issues. However, making the collected imagery data useful for automated defect detection requires significant manual labeling effort. Identifying assets and associated defects in image data requires creating an asset image labeling pipeline for subject matter experts.

We propose a novel solution that combines synthetic asset defect images with manually labeled drone images. This solution has several benefits: improves performance of defect detection, reduces the number of hours spent on manual labeling, and enables the capability to generate realistic images of rare defects where not enough real-world data is available. We employ a workflow that combines 3D modeling tools such as Maya and Unreal Engine to create photorealistic 3D models and 2D renderings of defective assets and their surroundings. These synthetic images are then integrated into our training pipeline augmenting the real data. This study implements an end-to-end Artificial Intelligence solution to detect assets and asset defects from the combined imagery repository. The unique contribution of this research lies in the application of advanced computer vision models and the generation of photorealistic 3D renderings of defective assets, aiming to transform the asset inspection process.

Our asset detection model has achieved an accuracy of 92 percent, demonstrating its effectiveness in identifying assets in drone imagery. We achieved a performance lift of 67 percent when introducing approximately 2,000 synthetic images of 2k resolution. In our tests, the defect detection model achieved an accuracy of 73 percent across two batches of images. Our analysis demonstrated that synthetic data can be successfully used in place of real-world manually labeled data to train computer defect detection model.

The use of drone imagery and automated process for defect detection reduces the amount of manual work while doing in-field inspections and significantly reduces the number of hours spent on manual labeling. This work lays the groundwork for future automated drone inspections, promising improvements in safety, cost-efficiency, and system resiliency. The study aligns with customer satisfaction goals by potentially leading to customer savings and increased satisfaction due to improved system reliability and resiliency. Moreover, this pilot will allow us to build out a drone imagery pipeline, potentially enabling future use cases around automated drone inspections.

## 1. INTRODUCTION

Traditional methods of asset inspection, characterized by manual, ground-based evaluations, have long been the cornerstone of maintaining electric power distribution systems. However, these conventional approaches are notably being time-consuming and exposing field crews to various hazards. Moreover, they often fall short in capturing the complete picture of asset health, particularly in hard-to-reach or visually obscured locations. In response to these challenges, the advent of drone technology has revolutionized the field of asset inspection. The integration of drone-based inspections amasses tens of thousands of aerial images through the comprehensive drone inspection program. This burgeoning repository of drone imagery presents a unique opportunity yet poses a significant challenge: how to efficiently process and extract actionable insights from this vast data trove.

The use of drones for asset inspection was investigated in several publications [1-9]. Work in [1] identifies asset and maintenance management as one of the areas with practical value in using unmanned aerial vehicles (UAVs) for inspections. In [2] a deep learning method based on Mask R-CNN is proposed for automated assessment of conditions of electrical towers. Work in [3] discusses how UAV together with thermographic analysis and Light Detection And Ranging (LiDAR) can be used to replace costly inspections performed by manned helicopter. In [4] it is emphasized how utilizing drone technology can significantly improve safety of inspections. Intelligent overhead line inspection based on UAV and supported by the hybrid network is presented in [5]. Study in [6] describes the system based on neural network that can automatically detect different faults and defects on electric power lines. Work in [7] demonstrates how UAV-based inspections can increase the efficiency and reduce cost of asset inspection by demonstrating that a single power line can be fully inspected in under 10 min using drone and one or two technicians. Work in [8] describes an end-to-end convolutional neural network (CNN) model for inspection of transmission power lines. Study in [9] describes the use case of drones with thermographic distance measurement used to inspect high and medium voltage overhead network.

Recognizing the potential to harness the power of artificial intelligence (AI) in this context, this paper explores an innovative solution—employing advanced computer vision and AI techniques to automate the detection of asset defects. We propose a method that not only enhances the performance of defect detection but also significantly reduces the manual labor involved in image review, enabling the generation of realistic synthetic images of rare defects, for which real-world data may be insufficient. However, the utility of this collected imagery is hampered by the need for extensive manual labeling—a bottleneck that our proposed solution aims to address.

In this paper, we delve into the development and implementation of an end-to-end AI solution for asset and defect detection. We discuss the creation of photorealistic 3D models and 2D renderings of defective assets using state-of-the-art 3D modeling tools and their integration into a comprehensive training pipeline. Through this study, we aim to demonstrate the transformative potential of AI in the realm of asset inspection, paving the way for more efficient, safer, and cost-effective utility management practices.

## 2. USE CASE BACKGROUND AND FORMULATION

The primary objective of this use case is to harness the potential of artificial intelligence and computer vision to revolutionize the asset inspection process in utility management. Specifically, this study aims to develop a highly efficient, AI-driven system for the automatic detection and identification of defects in electric power distribution assets, using a synergistic combination of real and synthetic image data. This endeavor seeks not only to enhance the accuracy and speed of defect detection but also to establish a scalable, AI-enabled framework that can be extended to various asset types across the utility spectrum, significantly reducing manual labor and improving overall grid reliability and safety. In this study we focus on use of drone collected images for asset management in distribution, with focus on detecting crossarms attached to the distribution poles and classifying them into defective and healthy instances. Examples of distribution crossarms are presented in Fig. 1. Examples of defective crossarms are presented in Fig. 2 a) for decay and b) for split.

We consider a system that has over 300,000 utility poles spread across the service territory, many of which are decades old. To keep these aging structures safe and reliable for our customers, each pole is inspected regularly. Traditionally, these inspections would be completed by vehicle or on foot. These methods were prone to treacherous terrain, line of sight, or vegetation making it a challenge to safely inspect the equipment. Fortunately, drones allow us to mitigate the safety concerns that come with traditional methods, while providing more detailed inspections. High quality images and video from multiple angles helps us to identify potential issues sooner, so they can be addressed earlier. Drones are also less disruptive to our customers. By flying drones directly to the targeted piece of equipment, it minimizes the need for crews to access private property, and significantly reduces the environmental impact of equipment inspections. In 2023, we inspected over 10,000 poles and continues to find new ways to embrace drone technology as the new status quo.

Described robust drone program has been instrumental in reducing in-field hazards for inspection crews by capturing tens of thousands of images to support their inspection programs. Despite the advanced nature of this drone program, most of these inspections still relied on human personnel to manually review the images and videos for defects. While drones have been proven to help mitigate hazardous terrain and field related safety concerns, utilities are not staffed to review 100,000 photos per year, weeding out healthy poles to identify defects. How could we use AI to limit the amount of data that would need to be reviewed by our industry Subject Matter Experts? A functional AI would challenge the status quo of traditional inspections, while providing actionable insights to help drive improvements to the safety and reliability of the electric grid. We could even provide a greater value for our customer by combining AI with a process flow to not only identify defects but to help optimize repairs. One of the primary challenges faced in this endeavor was the scarcity of labeled real-world data for every potential defect. Such data is crucial for training and testing AI models effectively. To circumvent this

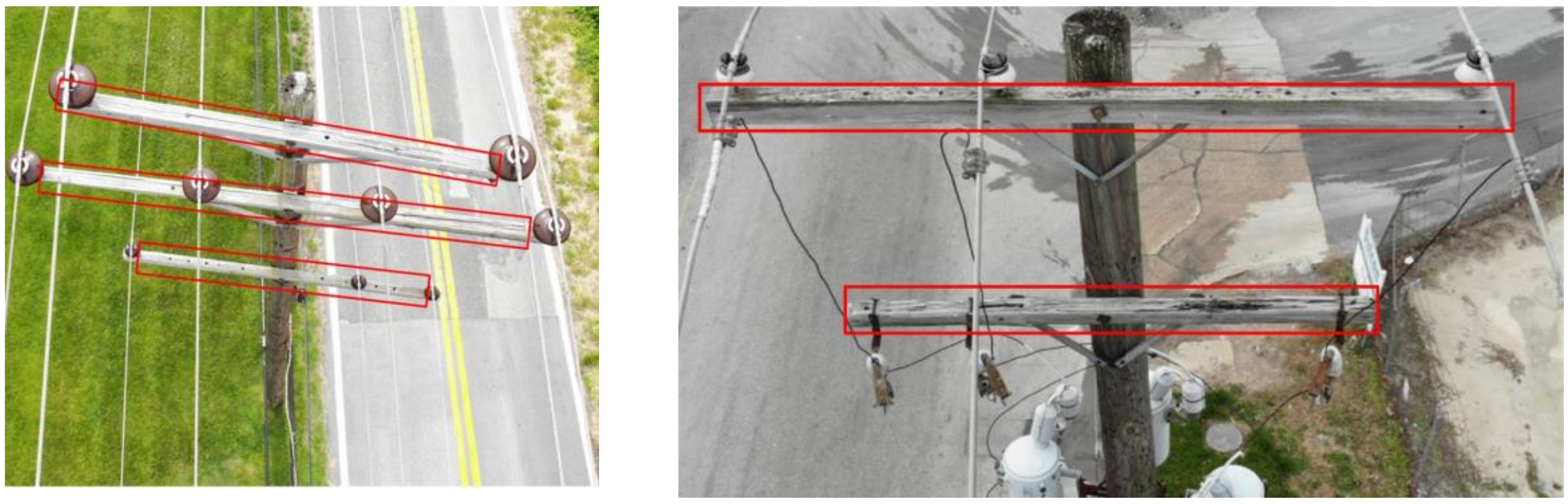

**Figure 1. Distribution crossarms**

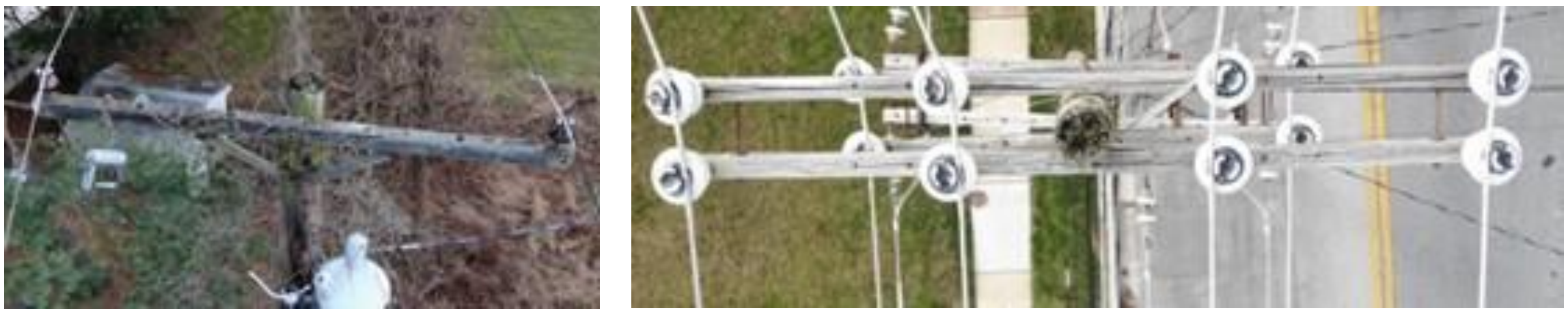

**a) Decay** **b) Split**

**Figure 2. Defective crossarms**

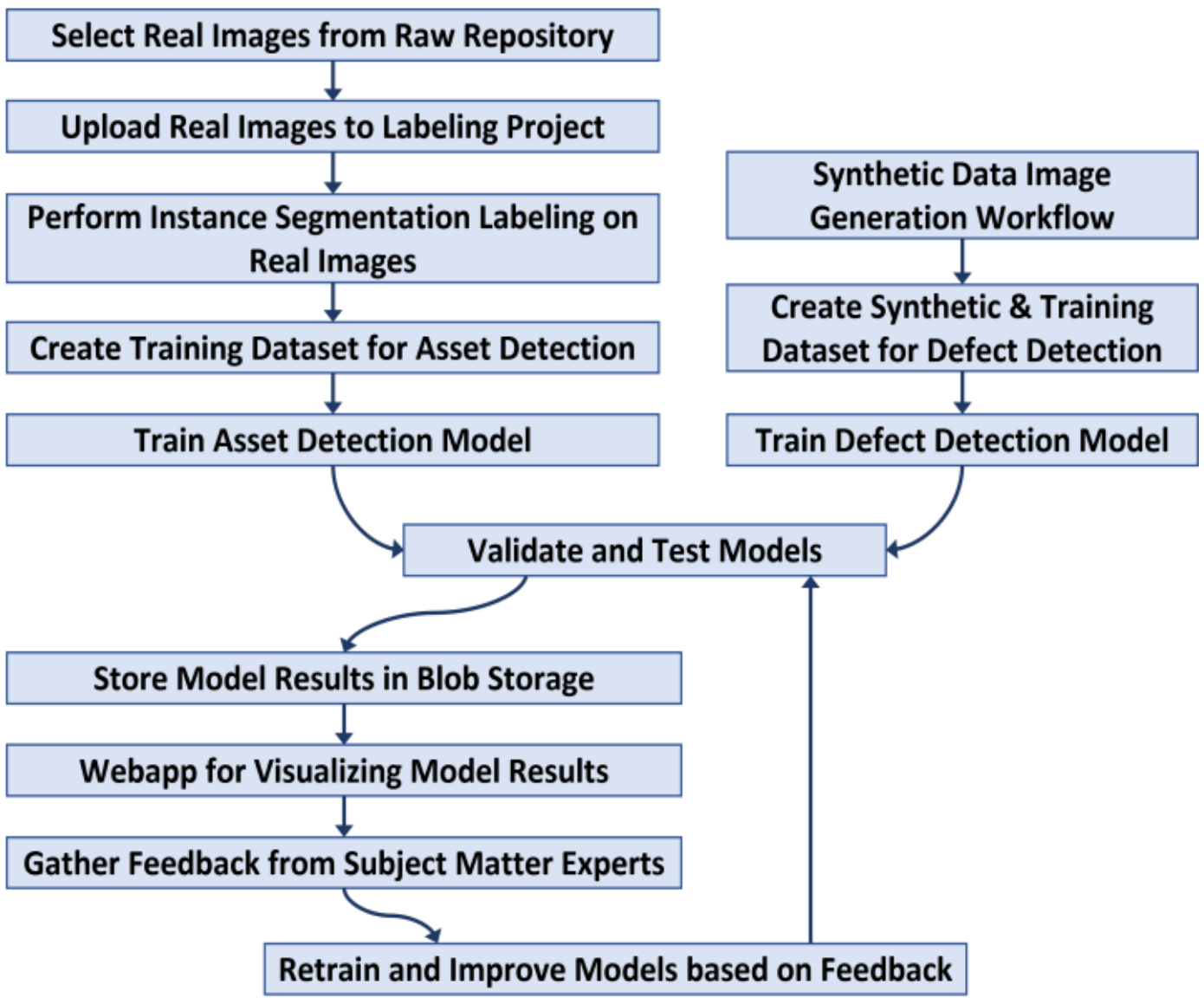

**Figure 3. Methodology Workflow**

issue, we leveraged synthetic data generation using Nvidia's Omniverse platform. Our goal was to create thousands of labeled, photorealistic examples of various defects, thereby enriching the dataset used for training the AI models.

## 3. METHODOLOGY

Our methodology encompasses a systematic workflow designed to integrate real and synthetic image data for training sophisticated AI models capable of identifying assets and defects. This process is iterative and involves several critical stages, Fig. 3:

- **Selection and Preparation of Real Images:** The initial step involves meticulously selecting relevant high-resolution images from the raw repository of drone-captured data. These images are then uploaded to a specialized labeling project platform tailored for this initiative.
- **Labeling and Annotation:** Using instance segmentation labeling, each real image is annotated to identify and delineate individual assets. This precise labeling process is essential to train the AI models with high accuracy.
- **Synthetic Data Generation Workflow:** In parallel, a synthetic data image generation workflow is employed. Utilizing advanced rendering and modeling software, this workflow generates photorealistic images of assets with various defects, which are then labeled accordingly.
- **Creation of Training Datasets:** Two distinct training datasets are created—one for asset detection and another for defect detection. These datasets are a blend of annotated real images and synthetic images, ensuring a comprehensive learning scope for the AI models.

**3.1. Model Architecture and Software Framework.** The research utilizes a two-tiered model architecture leveraging the capabilities of Detectron2, an open-source platform renowned for its efficiency in object detection tasks. The primary model employed is a Mask R-CNN, renowned for its precision in instance segmentation, which is critical for delineating the boundaries of assets and defects in utility images. The Mask R-CNN framework is particularly well-suited for this task due to its two-stage process, where it first proposes candidate object bounding boxes and then refines these proposals with high-quality segmentation masks. The architecture is anchored by a ResNet-101 backbone that provides a deep residual learning framework necessary for learning robust feature representations. Pre-trained on the COCO dataset, the ResNet-101 backbone brings transfer learning into play, capitalizing on learned features from a vast and varied dataset to improve the accuracy and speed of the defect detection process in the utility sector.

**3.2. Model Configurations.** The Mask R-CNN model was configured with specific hyperparameters tailored to the task at hand. The models were set to train with an image resolution of 2048 pixels to accommodate high-resolution drone imagery. The solver configurations were meticulously set up, with an initial learning rate of 0.001, employing a warm-up phase of 1000 iterations to prepare the model for the extensive learning phase that follows. The training was conducted over 50,000 iterations to allow the model to converge to a high level of accuracy, with checkpoints at every 1000 iterations to enable the model to resume training and avoid any loss of progress. To handle the extensive computational demands, the models were trained on NVIDIA Tesla T4 GPUs, which are optimized for machine learning workflows, offering a balance between performance and power efficiency. The GPUs were leveraged through CUDA 11.0, allowing the model to benefit from parallel processing for faster training times and more efficient data handling.

**3.3. Computational Resources.** The training and evaluation of our deep learning models were performed on a dedicated application server, equipped with state-of-the-art NVIDIA Tesla T4 GPUs.

These GPUs are specifically designed for machine learning and data center workloads, offering the necessary computational power to process our extensive datasets efficiently. With CUDA version 11.0 support, these GPUs provide a robust platform for running our PyTorch-based models, enabling rapid iterations and evaluations. The server's capacity, indicated by the substantial memory allocation per GPU, ensures that our models can be trained on large batch sizes and complex network architectures, which is vital for achieving high accuracy in asset & Defect detection.

**3.4. Training Pipeline.** The Detectron2 framework was utilized to construct a robust training pipeline. A custom training loop was scripted using Python, with batch sizes and other configurations set to optimize GPU utilization. The model weights, initialized from the pre-trained Mask R-CNN, were fine-tuned with the utility-specific dataset comprising both real and synthetic images. During the training phase, data augmentation strategies such as random flips, brightness adjustments, and cropping were employed to increase the diversity of the training data, which helps the model generalize better to new, unseen images. The models underwent continuous evaluation against a validation set to ensure that performance was consistently improving and to prevent overfitting, in several steps:

- **Validation and Testing of Models:** Post-training, both models undergo a rigorous validation and testing phase. This step ensures that the models perform with high accuracy and reliability, meeting the predefined standards for defect detection.
- **Storing and Accessing Model Results:** The results, including the accuracy metrics and detection outputs of the models, are stored securely in blob storage. This centralized storage facilitates easy access and management of the model data.
- **Visualization of Results:** A dedicated web application is developed for visualizing the model results. This platform is an interface for stakeholders to interact with the findings of the AI models.
- **Expert Feedback Integration:** Subject matter experts provide feedback on the model's performance based on real-world applicability and accuracy. Their insights are crucial for contextualizing model results and identifying areas for improvement.
- **Model Retraining and Refinement:** Leveraging the feedback received, the models are retrained and refined to enhance their accuracy and utility. This iterative process of retraining ensures continuous improvement and adaptability of the AI models.

## 4. SYNTHETIC IMAGE CREATION AND WORKFLOW

In this section we describe the process of 3D Modeling of Defects. We develop precise 3D models of three critical defects—cross-arm splits, breaks, and decay. These models were designed to capture the intricate details of the defects, providing the AI with high-quality representations for learning. The overview of the development of photorealistic synthetic images is presented in Fig. 4.

**4.1. Randomization and Variability.** To simulate the complexities of the real world, we introduced randomization across several dimensions:

- **Cosmetic Variations:** Mimicking the wear and tear and individual peculiarities found in real-world.
- **Camera Angles:** Diversifying the perspectives from which assets and defects are viewed, as they would be in drone-captured images.
- **Lighting and Environmental Conditions:** Ensuring models are exposed to a range of lighting scenarios and environmental contexts, such as different times of day and weather conditions, to enhance the AI's ability to generalize from the synthetic data to real-world situations.
- **Inclusion of Distractors:** To prevent the AI models from overfitting to too-clean scenarios, up to 50 generic distractor objects were introduced into the synthetic scenes. These objects serve to replicate the cluttered and unpredictable nature of actual environments.

**4.2. Comprehensive Labeling.** Each image was labeled for multiple purposes:

- **Classification:** Categorizing the type of defect or asset present in the image.
- **Object Detection:** Marking the bounding boxes around each asset and defect.
- **Semantic Segmentation:** Labeling each pixel of the image for all main objects to understand the context and shape of the assets and defects fully.

**4.3. Volume and Quality of Images.** To ensure the robustness of the model, a large volume of images was deemed necessary. A total of 10,000 images were generated for each type of defect, culminating in a dataset sizeable enough to train the AI effectively. Moreover, to maintain the fidelity required for accurate AI analysis, each image was rendered at resolutions up to 4k. To augment our dataset with high-fidelity synthetic images, we employed a sophisticated workflow that leverages the strengths of several cutting-edge tools and platforms. The process begins with sourcing 3D assets which are either existing client assets or derived from 3D content factories and third-party content providers. These assets serve as the foundational elements for creating realistic utility infrastructure models.

**4.4. Asset Development and Scene Creation.** The assets are imported into Maya, a powerful 3D modeling and animation software, where they undergo further development. This includes detailed modeling and material design to ensure that the assets accurately reflect real-world textures and lighting conditions. Connectors, acting as integrative links, then feed these enhanced assets into the Omniverse Nucleus database. Omniverse Nucleus serves as the central hub for collaboration, enabling various applications to work together seamlessly. It allows for the real-time updating and sharing of scene descriptions, making it an essential component of the synthetic data generation pipeline.

**4.5. Procedural Scene Design and Randomization.** With the assets in the Omniverse Nucleus, we utilize Houdini's procedural scene design tools to generate complex scenes and environments. These scenes are designed to mimic the real-world settings in which the utility assets are situated, accounting for a variety of geographical and environmental factors. Further complexity is added using NVIDIA's graph creation tools, which introduce randomization in asset placement and defect characteristics. This randomness is key to creating a diverse set of images that a model can learn from, ensuring that the AI is not just accurate but robust against a wide range of scenarios.

**4.6. Resultant Synthetic Images.** The results of this workflow are high-quality, photorealistic images that significantly enhance the training dataset for our AI models. These images, exemplified in the diagram, showcase utility assets in realistic settings with varying defect presentations. The synthetic images are designed to be virtually indistinguishable from real drone-captured images, thus providing the AI with a comprehensive understanding of the assets and potential defects. Example of real vs. synthetic image can be viewed in Fig. 5.

## 5. ASSET DETECTION

Computer vision allows computer systems to identify, classify, and understand a specific instance (object) in images and videos. Some of the main computer vision methods include image

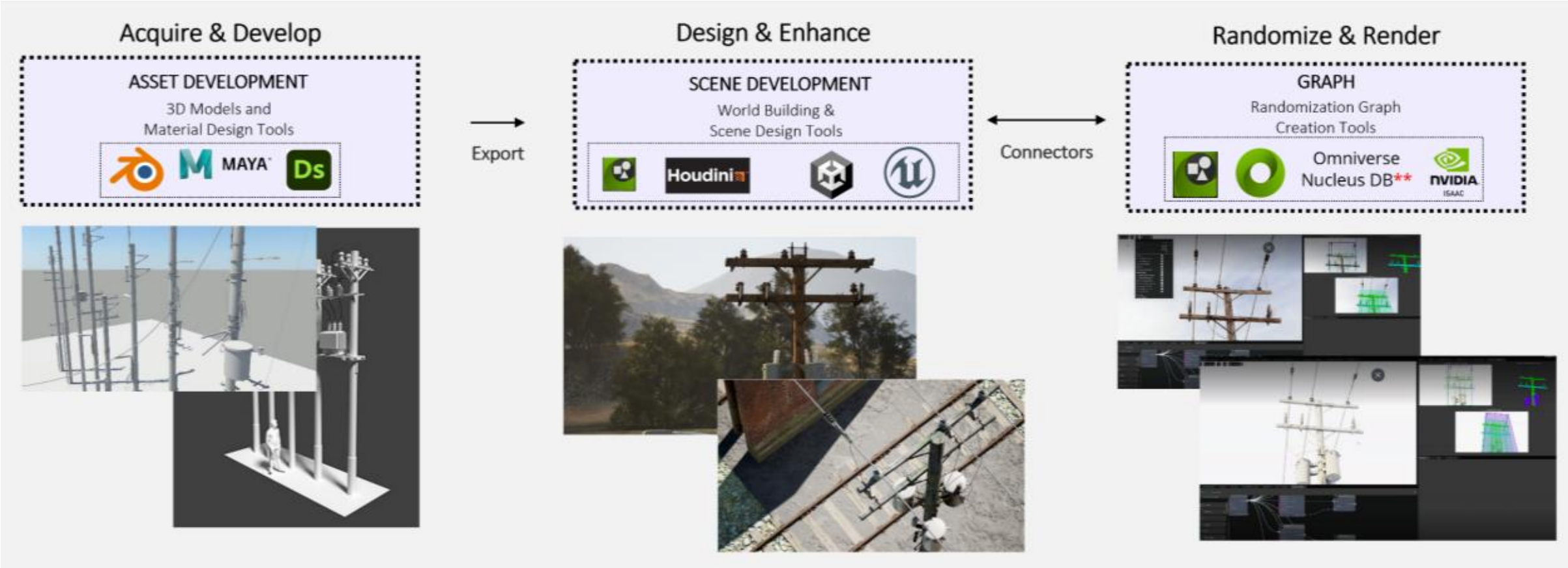

**Figure 4. Development of photorealistic synthetic images [10]**

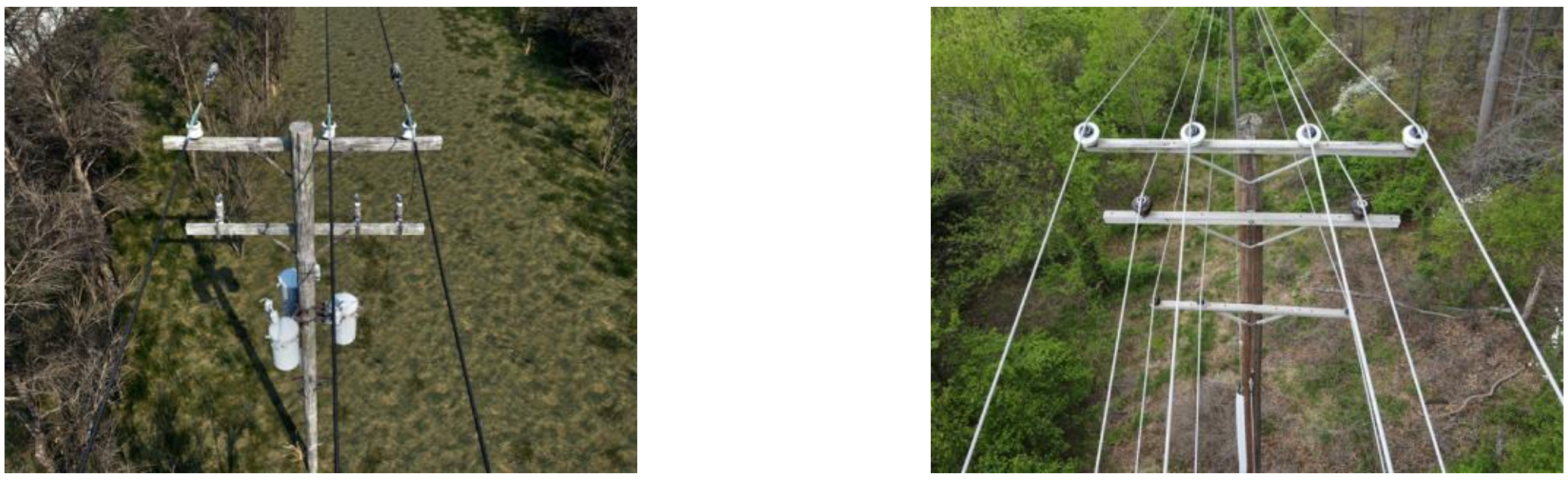

**a) Synthetic Image** **b) Real Image**

**Figure 5. Comparison of synthetic vs. real image**

recognition, object detection, semantic segmentation, and instance segmentation, Fig. 6. In this study we use instance segmentation to detect crossarms in drone-collected asset inspection images. This section of our study delineates the methodical approach we undertook to construct a robust dataset for training our asset detection model, an endeavor critical for the accurate identification of utility assets from drone-captured imagery. Our initial step involved a thorough data exploration phase, where we consolidated data from the existing repository. This phase was crucial to understanding the variety and distribution of assets within our dataset. We performed a high-level summary of crossarm defects, ensuring we had a representative sample of various defect conditions within our dataset. Following procedure was used for labeling:

- **Crossarm Labeling Protocol:** Prior to the labeling process, we established a rigorous crossarm labeling protocol. This protocol was carefully designed to maintain consistency across the dataset, ensuring that the crossarms were labeled uniformly. By employing instance segmentation, we accurately marked the boundaries of each crossarm, thus capturing their true shape and size. This meticulous process allows our model to learn from precise geometric representations of the assets.
- **Crossarm Labeling Effort:** We embarked on a substantial labeling effort, leveraging Microsoft Azure Data Labeling Studio for its robust toolset that met our labeling requirements. The platform enabled us to label large volumes of images efficiently, with features that supported our instance segmentation needs.
- **Label Data Preprocessing:** Post-labeling, we engaged in a comprehensive label data preprocessing phase. This involved formatting the data, such as converting relative coordinates to absolute coordinates and filling up image width and height details in the COCO file format, widely used for object detection and instance segmentation algorithms.

## 6. DEFECT DETECTION ENHANCED WITH SYNTHETIC IMAGES

Building upon the established processes for the asset detection dataset, the creation of the defect detection dataset introduced a crucial additional step: the integration of synthetic images, Fig. 7. This integration was meticulously planned to ensure the synthetic images enhanced the learning process without introducing biases.

Upon generation, synthetic images underwent a rigorous validation process to ensure their suitability for training the defect detection model. This process was critical in maintaining the integrity of the training dataset and ensuring the model would not learn from inaccurate or unrealistic data. This process served two primary purposes. Firstly, to ensure that our asset detection models can identify objects within these synthetic scenarios. Secondly, to evaluate the realism and applicability of these images for subsequent defect detection.

As part of our initial object detection validation, we commenced the validation process by running inference with our pre-trained asset detection models on the synthetic images. This initial test aimed to verify whether the models could detect objects in these images. Successful object detection is a prerequisite before any defect detection can be considered, as the absence of detectable objects would render the synthetic images ineffective for further analysis.

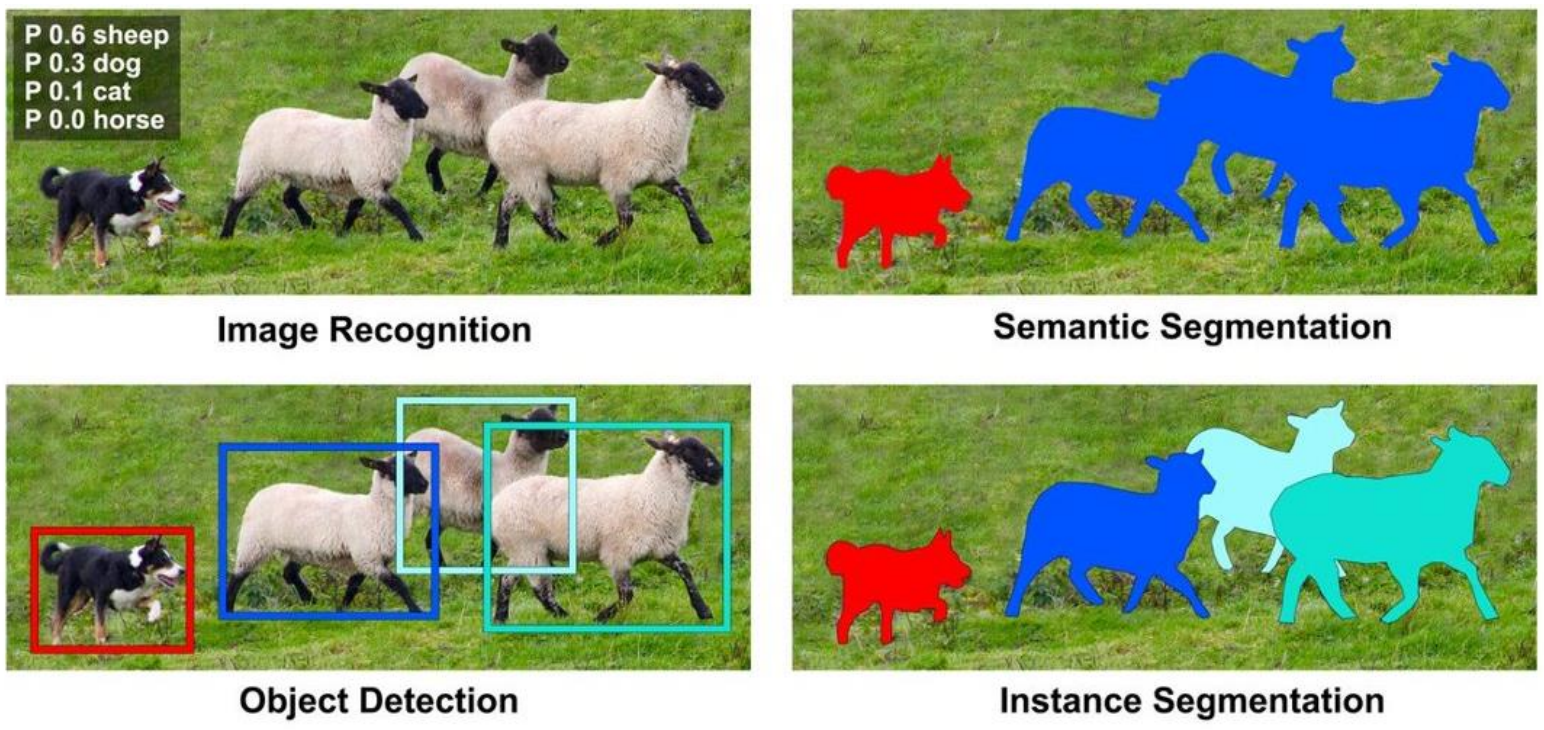

**Figure 6. Computer vision methods**

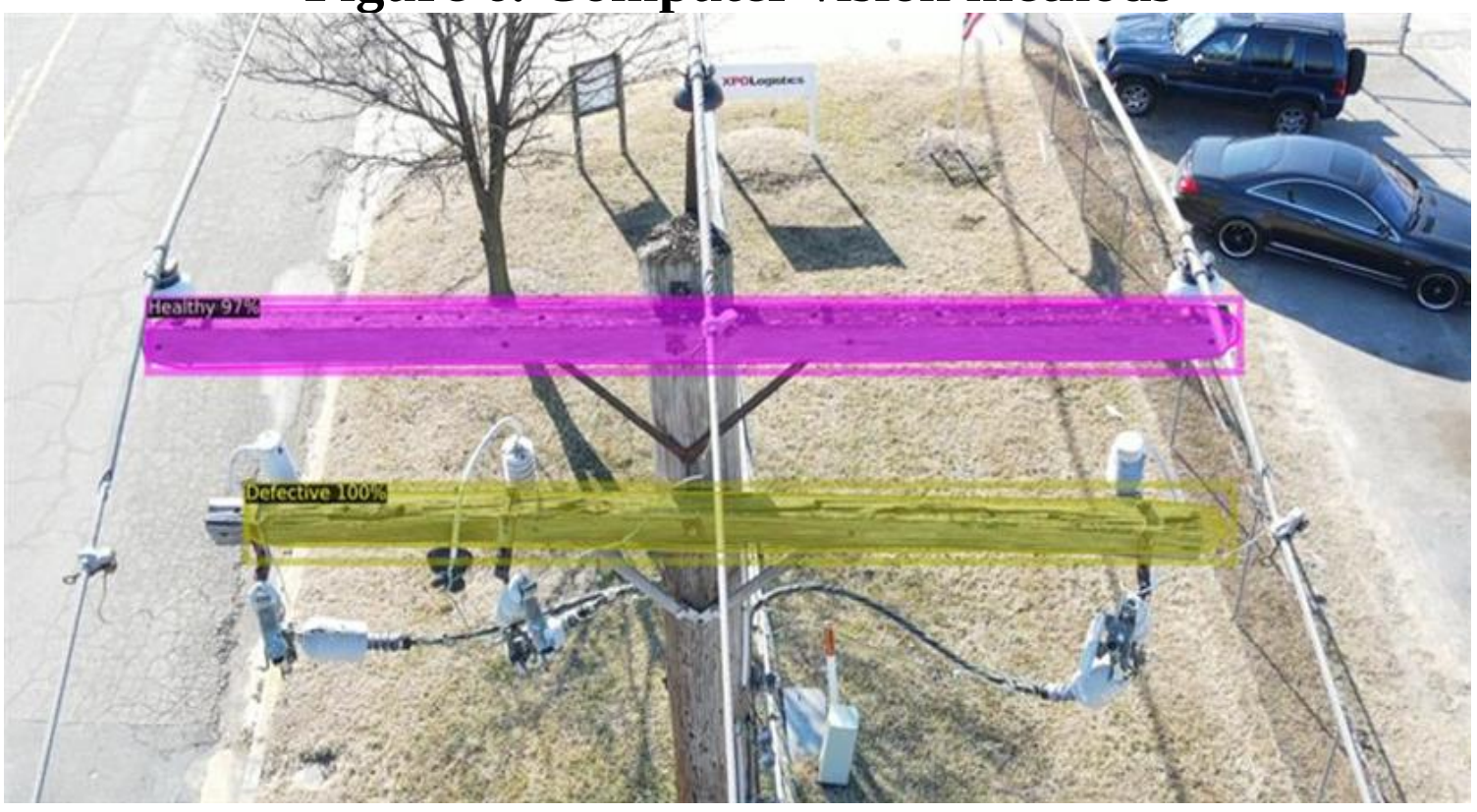

**Figure 7. Asset and Defect Detection: Cross-Arms with Identification**

Following successful object detection, we closely examined the synthetic images for realism. This involved comparing the detected objects against their real-world counterparts to ensure the synthetic images were representative and could be used for practical training and evaluation of the models. This step was essential to guarantee that the synthetic images would not introduce unrealistic bias into the model training process.

The error analysis process played a pivotal role in refining the synthetic images. Any discrepancies in object detection were meticulously analyzed, and the synthetic image generation parameters were adjusted accordingly. This iterative process helped in fine-tuning the synthetic data to closely mimic real-world conditions, thereby enhancing the efficacy of the subsequent defect detection.

Armed with insights from the initial object detection and error analysis, we focused on expanding the set of synthetic images that demonstrated high effectiveness in being detected by the asset detection models. This approach was data-driven and ensured that the efforts were concentrated on generating more of the synthetic images that provided the most value for our models.

This process of validation, error analysis, and refinement of synthetic images was not a one-off task but an ongoing endeavor. As the asset and defect detection models evolved, the synthetic images were continuously assessed and improved to maintain their relevance and usefulness in training robust, real-world applicable AI models.

We implemented several quality control measures, including:

- **Annotation Review:** Each synthetic image was reviewed to ensure that the annotations accurately reflected the intended defects.
- **Defect Distribution Analysis:** We analyzed the distribution of defects across the synthetic images, ensuring a representative variety of defect types and severities that mirrored real-world conditions.
- **Balance Between Healthy and Defective Samples:** Ensuring a balance between healthy and defective samples within the dataset to teach the model to identify the presence or absence of defects.
- **Categorical Distribution:** We ensured that the categories of defects were proportionately represented according to their prevalence in the real-world, as indicated by our initial exploration.
- **Dataset Validation:** Before proceeding to model training, the integrated dataset underwent a validation process where a subset of the data was reviewed by SMEs to confirm the accuracy of defect labels and the realism of synthetic images in depicting true defects.
- **Final Dataset Composition:** The final dataset composition was carefully documented, detailing the total number of images, the breakdown of annotations by defect type, and the healthy versus defective image ratio. This documentation provides transparency and allows for replicability.

## 7. END-TO-END IMAGE TRACKING AND VERIFICATION PIPELINE

To support the continuous integration of asset and defect detection from drone images with the asset management, the end-to-end pipeline is developed that automates the following steps that occur throughout the model lifetime, as presented in Fig. 8:

- Incoming raw drone collected images are automatically detected from the Azure BLOB storage and distribute between the Batch Prediction and Labeling.
- All the images in Batch Prediction are automatically ran through the Inference Model.
- After the inference is completed, results of classification are automatically sent to the Web application to be verified by the SMEs. The system uses WebApp verification table to track the progress and results of verification. Examples of are presented in Fig. 9 in Map View and Fig. 10 in Table View.
- Based on the results from the WebApp verification, images with incorrect prediction are automatically sent to Staging.

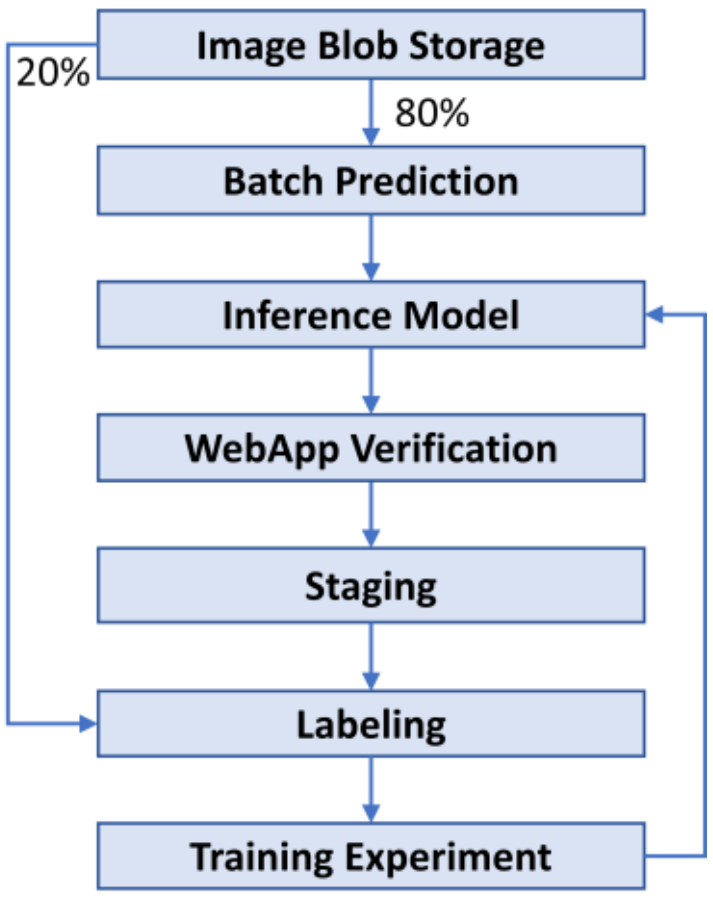

**Figure 8. End-to-end image tracking pipeline**

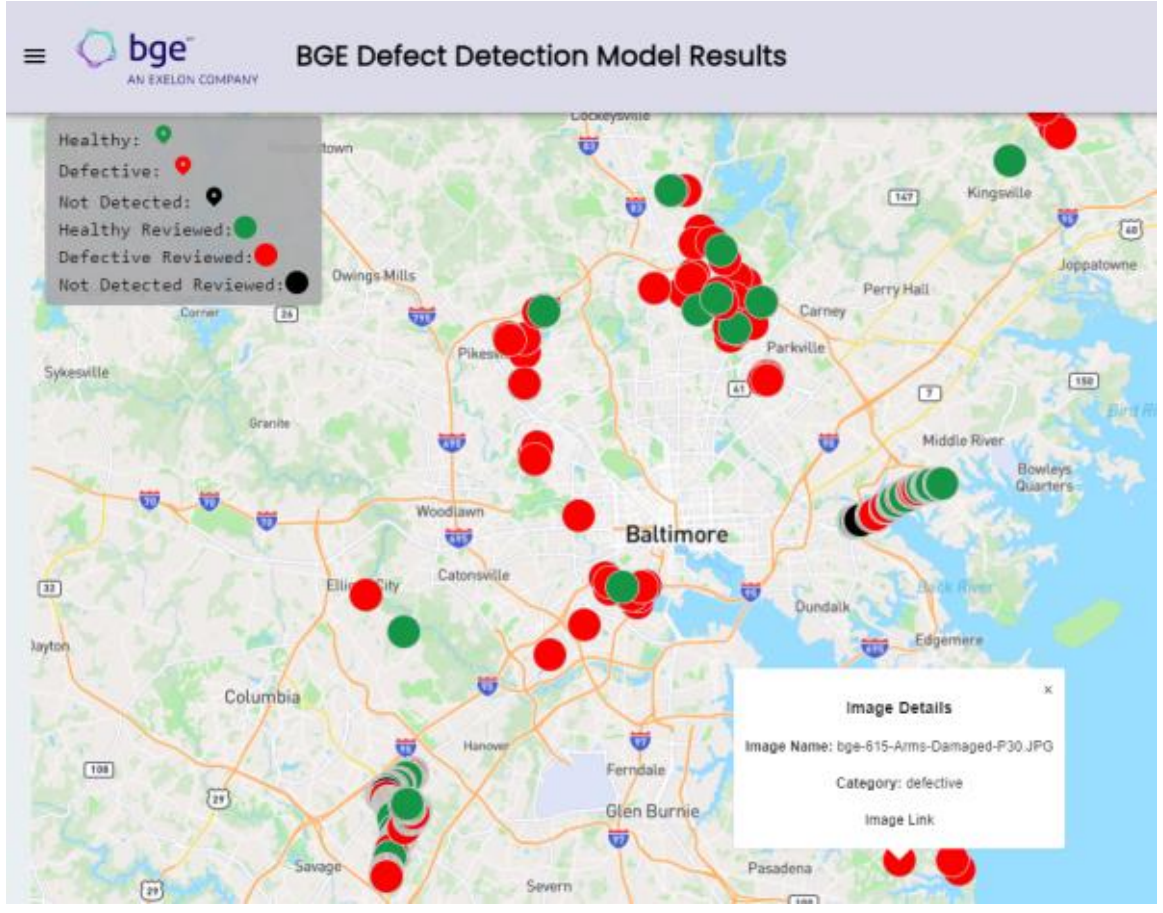

**Figure 9. WebApp Verification Platform – Map View**

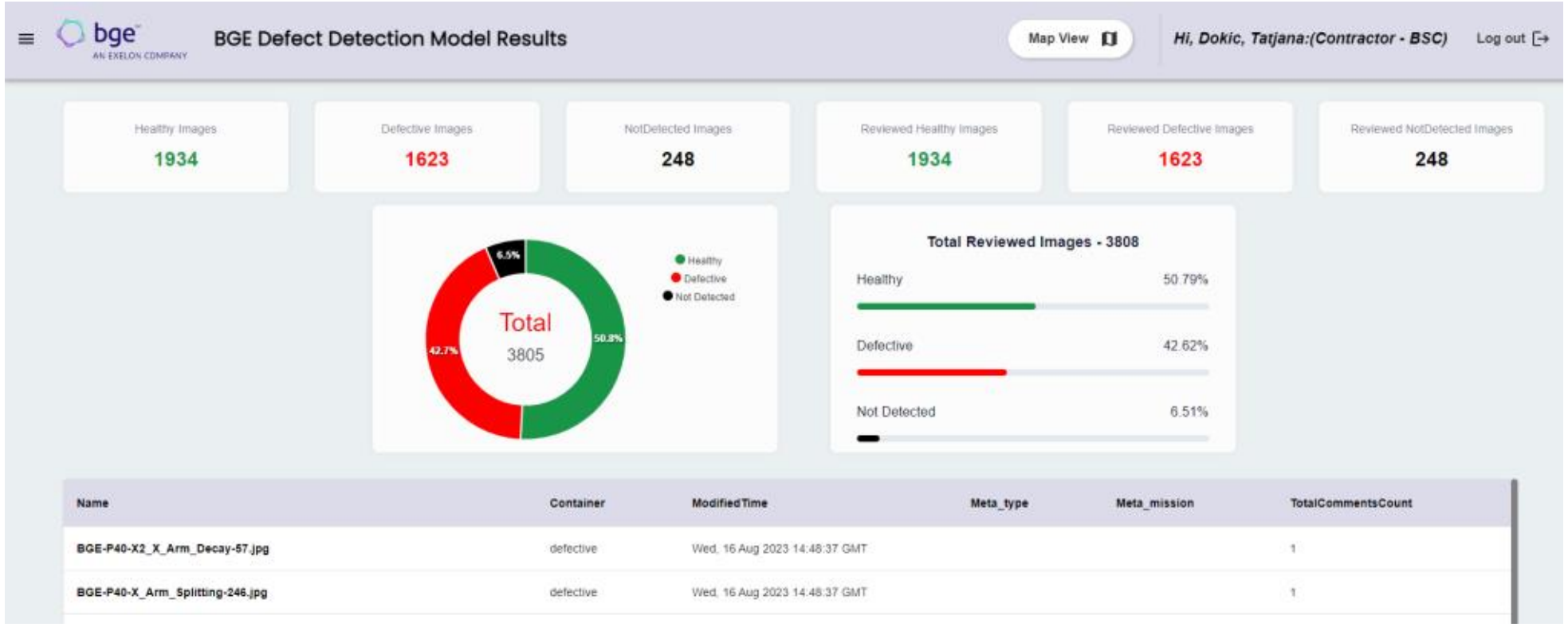

**Figure 10. WebApp Verification Platform – Table View**

- Selected image from the Staging together with any other images marked for Labeling are loaded to the Labeling Platform. After labeling is complete, the images with their assigned labels are automatically sent to the Training Experiment.
- The labeled images are repeatedly used in Training, improving the model with every iteration.
- Image Tracking system was implemented using Azure Table that tracks the entire lifetime of each image as it transitions from one state to another inside the model pipeline presented in Fig. 8.

## 8. RESULTS

**8.1. Asset Detection Model Training and Evaluation.** The asset detection model underwent a comprehensive training and evaluation process over four iterations. This iterative approach allowed for continuous refinement and improvement of the model's performance. For each iteration, new images were added to the training pool to enhance the model's ability to generalize across a wider array of scenarios. By the third iteration, the training set was enriched with a total of 517 high resolution images providing a diverse range of examples for the model to learn from.

The evaluation of the model was conducted using a separate test set comprising 1044 images. These images were carefully selected by individual users to ensure a wide representation of inference scenarios. The model was trained using a curriculum that progressively introduced more complex images and annotations. The curriculum was designed to first establish foundational detection capabilities and then incrementally introduce challenges, such as varying defect types and environmental conditions. The model's performance was quantified using precision, recall, and the F1-score, which is the harmonic mean of precision and recall. These metrics were chosen for their ability to provide a balanced view of the model's accuracy in identifying assets and defects.

The model was trained using Detectron2, a versatile open-source library for object detection. The training pipeline was scripted in Python, with a focus on utilizing GPU acceleration to handle the computation-intensive training process efficiently. The dataset, consisting of images and annotations, was prepared in a directory structure compatible with Detectron2's requirements. Various configurations were set, such as the number of workers, batch size, learning rate, and model weights, which were sourced from a pre-trained Mask R-CNN model. Custom functions were defined to build evaluators and data loaders, incorporating potential data augmentation strategies for a robust training regime.

The configuration details and the custom trainer class were vital in fine-tuning the model to the specific task of asset detection. The model was configured to predict segmentation masks, an essential feature for precise asset boundary detection. Transfer learning was leveraged by importing pre-trained weights, a strategy that significantly improves training efficiency and model performance. The model underwent 40,000 iterations of training with periodic evaluations and checkpoints, ensuring thorough learning and the opportunity for continuous assessment.

*Results*. As presented in Fig. 11 the F1-scores across the iterations showed a consistent improvement, with the third iteration achieving a score of 0.92, indicating high precision and recall. The number of annotations and training images used in each iteration were documented, showcasing the scale of data that the model was exposed to. A bar chart was provided in Fig. to visually represent the F1-scores and the number of annotations in each iteration. This visual aid effectively communicates the progression and success of the model training across iterations.

**8.2. Defect Detection Model Training Curriculum and Iterations.** Our defect detection model underwent several experiments to identify the optimal balance between real and synthetic images. Each experiment iterated on the previous, gradually incorporating more synthetic images to assess their impact on model performance.

- **Baseline:** Established with real images only, setting a performance benchmark.
- **Experiment 1:** Introduced synthetic images equal to the number of real images.
- **Experiment 2:** Doubled the synthetic images to further test model robustness.
- **Experiment 3:** Increased the synthetic images fivefold, analyzing the scalability of model.
- **Experiment 4:** Maintained the high volume of synthetic images while increasing the resolution, aiming to enhance model precision further.

Our model selection was driven by the goal of achieving high accuracy in defect detection on various utility assets. We leveraged the capabilities of the state-of-the-art architecture provided by the Detectron2 framework, utilizing a Mask R-CNN with a ResNet-101 backbone pre-trained on COCO dataset weights. This choice was informed by the model's proven efficiency in handling complex image segmentation and object detection tasks.

To validate the effectiveness of our models, we employed the Mean Average Precision (MAP) at different Intersection over Union (IoU) thresholds. Precision at IoU thresholds of 0.5 and 0.75 and the mean average precision over a range from 0.5 to 0.95 were considered. Confidence thresholds and maximum detections per image were fine-tuned to optimize the F1 score, using a validation set to estimate the optimal threshold for decision-making, Fig. 12. A confidence threshold of 0.9 was used for all evaluations. The optimal confidence threshold that maximizes a given metric (like F1) on a test set might be different. An analysis against a validation set with similar distribution as the test set can be used to obtain an estimation of the optimal threshold as seen in the Fig. 12. At inference or evaluation time, extra detections with lower confidence can be suppressed using this heuristics if for instance no more than 6 cross-arms are expected to be present in any one image. This will be tied to the data collection strategy during inference time. This parameter has not been set for comparison purposes with baseline but is recommended during inference.

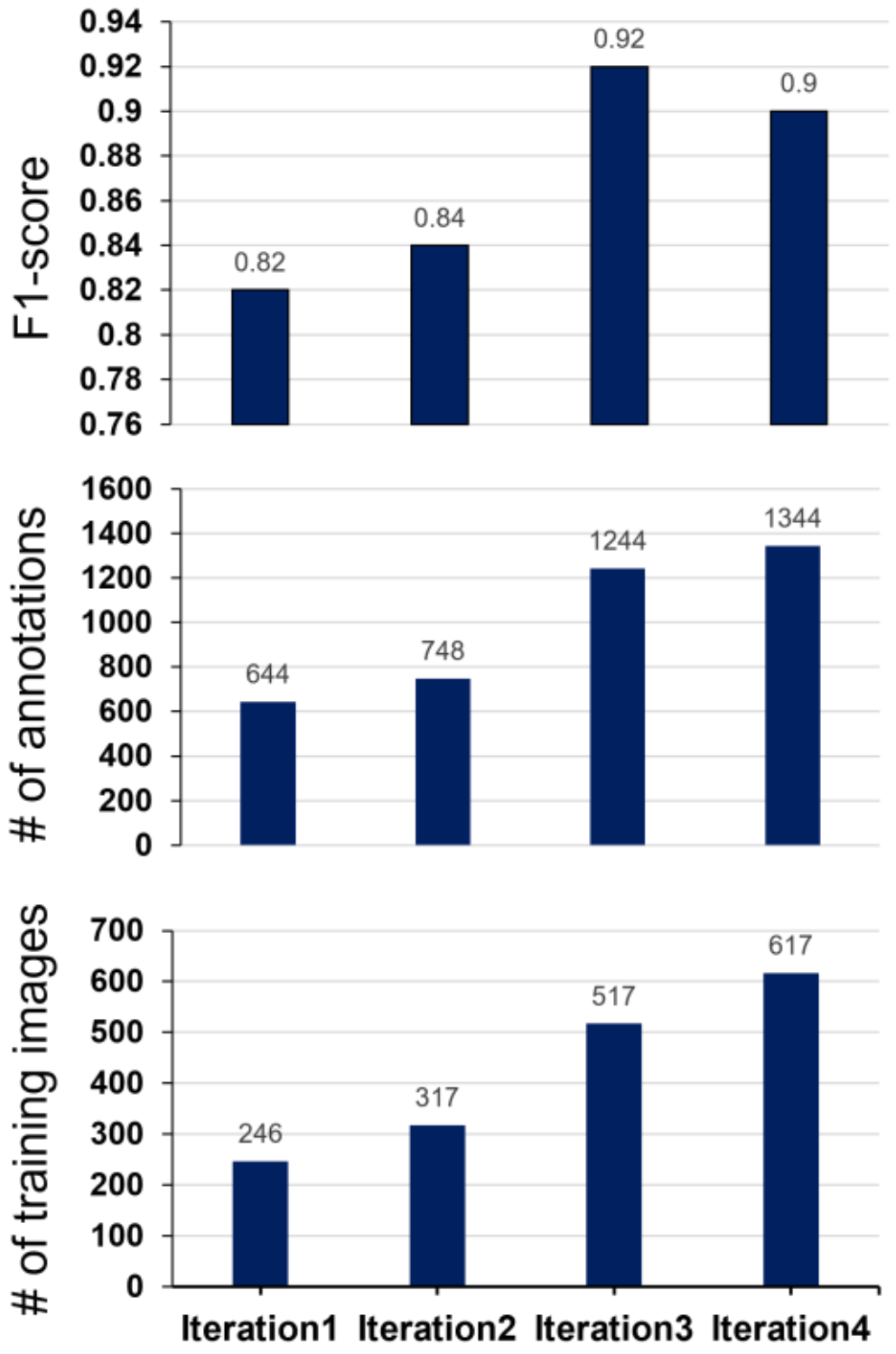

**Figure 11. Asset Detection Model Results**

*Results*. The model was tested on 3805 images containing 10230 labelled crossarms (instances). Confusion matrix on instance level is presented in Fig. 13. Performance metrics for classifying instances into healthy and defective is presented in Table I. The performances were calculated using the equations:

$$Precission_{healthy} = \frac{True\ Healthy}{True\ Healthy + False\ Healthy},\ Precission_{defective} = \frac{True\ Defective}{True\ Defective + False\ Defective}$$

$$Recall_{healthy} = \frac{True\ Healthy}{True\ Healthy + False\ Defective},\quad Recall_{defective} = \frac{True\ Defective}{True\ Defective + False\ Healthy}$$

**8.3. Discussion.** The experiments revealed a clear trend: the integration of synthetic images improved model performance, with Experiment 4 showcasing a significant increase in MAP. The optimal threshold for confidence was determined to be 0.9, balancing precision and recall effectively. Our model demonstrated excellent precision at higher IoU thresholds, indicating a strong ability to distinguish between healthy and defective assets accurately.
The F1-Confidence curve analysis further solidified our confidence threshold choice, ensuring that our model performed optimally when identifying defects across a variety of utility assets. The inclusion of a large set of high-quality synthetic images was instrumental in achieving a performance lift in MAP of 66.96%, a notable improvement from the baseline model, as can be seen in Table II.

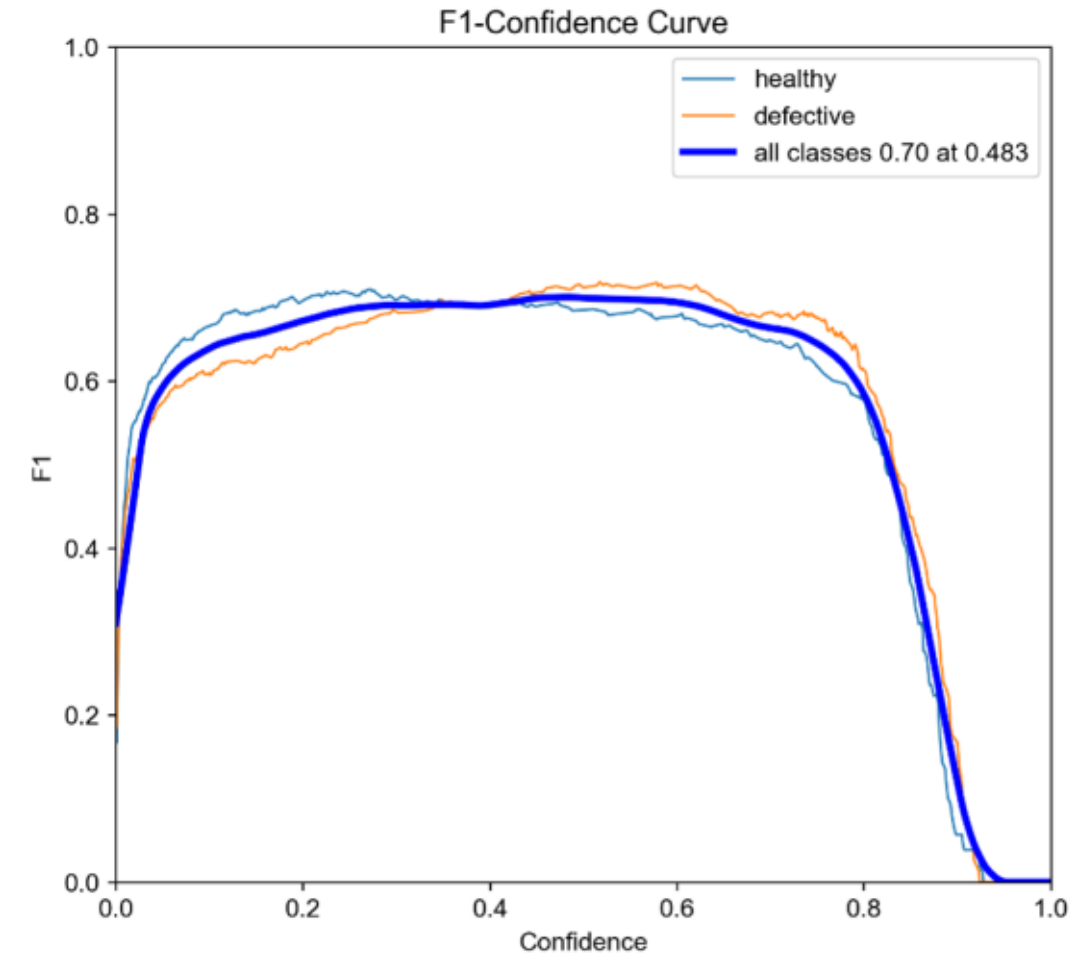

**Figure 12. Example of a F1 vs. Confidence chart to determine optimal confidence by analysis on validation set.**

## 9. CONCLUSIONS

In this study we propose combining synthetic asset defect images with drone-collected images to achieve higher performances of the asset defect detection model for distribution crossarms. The implementation of the AI-driven system for asset and defect detection, as detailed in this paper, presents several tangible benefits that extend beyond the technical sphere into operational efficiency and business gains. These include:

- The use of drone imagery and automated process for defect detection reduces the amount of manual work while doing in-field inspections and significantly reduces the number of hours spent on manual labeling.
- We propose an effective model for asset detection from drone images with accuracy of 92%. We demonstrate that our defect detection model has Precision of 95.89% and Recall of 89.07% for healthy images, and Precision of 58.12% and Recall of 79.87% for defective images.
- Including synthetic images improves performance of defect detection, while significantly reducing the time needed for manual labeling. With the inclusion of synthetic images, we achieved a performance improvement of 67 percent. With this result we confirm that synthetic data can be efficiently used to complement real-world drone images when training computer vision model.

Despite the significant advancements in utilizing AI for drone-based asset inspection, our current process

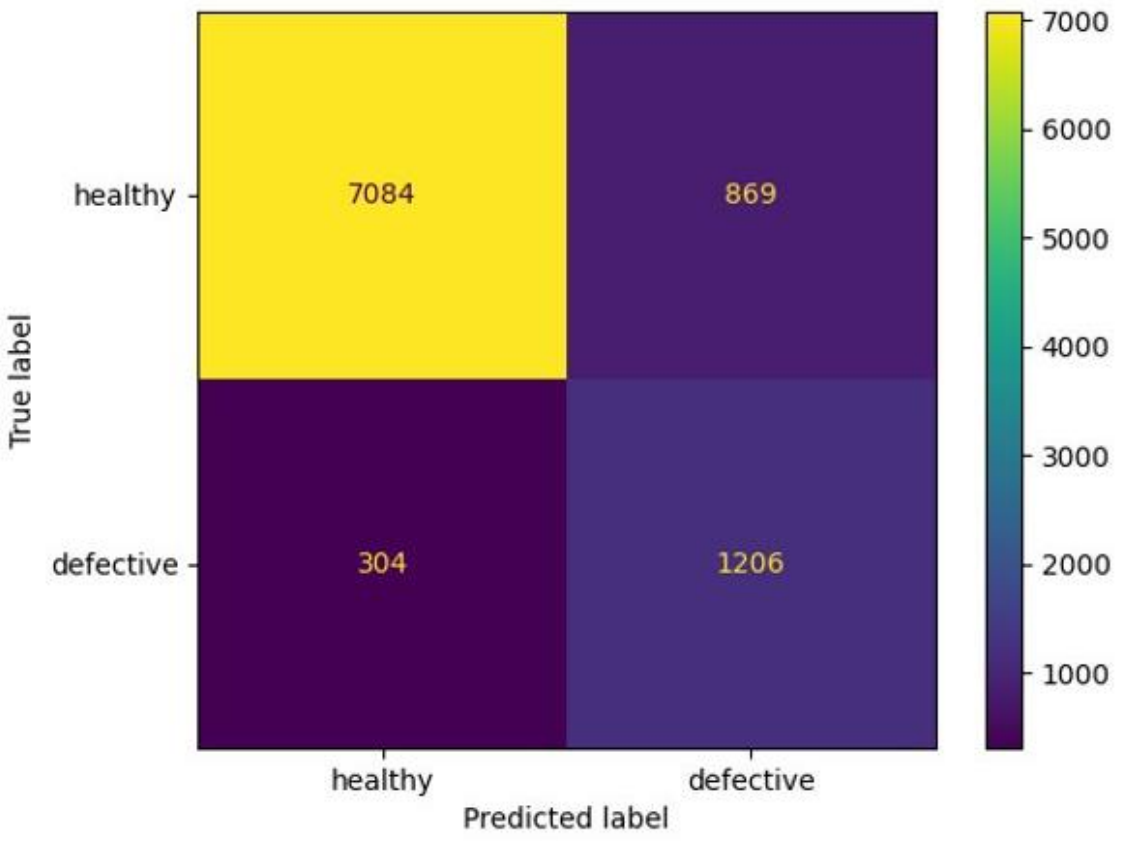

**Image 13. Instance Confusion Matrix**

**Table I. Model Performance**

| Class | Precision | Recall |
|---|---|---|
| **Healthy** | 95.89% | 89.07% |
| **Defective** | 58.12% | 79.87% |

**Table II: Defect Detection Model: Augmentation with Synthetic Images**

| Model | Training Images (Real) | Training Images (Synthetic) | Training Image Resolution | Mean Average Precision | % Increase from Baseline |
|---|---|---|---|---|---|
| Baseline | 393 | 0 | 2k | 20.58 | N/A |
| Exp 1 | 393 | 393 | 1k | 25.64 | 24.60% |
| Exp 2 | 393 | 786 | 1k | 27.22 | 32.27% |
| Exp 3 | 393 | 1,965 | 1k | 27.98 | 35.96% |
| Exp 4 | 393 | 1,965 | 2k | 34.36 | **66.96%** |

is not without its limitations. The most notable of these is the substantial time requirement for manual real image labeling, which remains a critical bottleneck. Manually annotating images for training our detection models is labor-intensive, prone to human error, and not easily scalable, potentially impacting the speed and efficiency of deploying AI solutions for asset and defect identification. In the future we aim to improve our processes by following steps:

- **Incorporation of SAM** (Segment Anything Model) into our existing asset & defect detection models SAM's ability to identify and segment any object within an image without the need for prior training or assigned labels presents a breakthrough in automating the labeling process.
- **Leverage Existing Models**: Integrate SAM with our existing asset detection models to automatically extract bounding boxes and apply class labels, thus streamlining the preparation of training datasets.
- **Automate the Labeling Process**: Reduce the project development time traditionally devoted to image labeling and annotation, particularly for instance segmentation problems.
- **Enhance Data Consistency**: Improve the consistency and accuracy of labeling across the dataset, which is critical for training reliable AI models, and increase the scalability of our data processing pipeline, allowing for more extensive datasets and faster iterations in the model development cycle.
- **Training and Deploying Diverse Models:** We plan to train and deploy at least 100 varieties of asset and defect detection models across all opcos. This diversity is necessary to account for the different asset types, environmental conditions, and operational requirements unique to each service territory. By doing so, we aim to create a comprehensive AI-driven inspection framework that is adaptable and robust across multiple contexts.
- **Upgrading Computational Infrastructure:** To support the increased computational demands of training and deploying numerous models, we are scaling up our GPU infrastructure from NVIDIA Tesla T4 GPUs to the more advanced H100 GPUs. This upgrade will enhance our computational capabilities, improving our vision model training times by up to 4x compared to the existing T4 GPUs. The increased processing power will facilitate both batch and real-time training and inference needs, enabling us to handle larger datasets and more complex models efficiently.
- **Integration into Business Applications:** We are focusing on integrating all these models into downstream business applications. Embedding the AI models into our operational workflows will provide faster asset and defect analysis for crew engineers, enabling them to prioritize work orders more effectively. This integration is crucial for translating technological advancements into tangible operational improvements and cost savings.